%% file: paper.tex
\documentclass[10pt,twocolumn,letterpaper]{article}

\usepackage{iccv}

\input{vcl-shortcuts.tex}

\usepackage[pagebackref=true,breaklinks=true,letterpaper=true,colorlinks,bookmarks=false]{hyperref}

\iccvfinalcopy 

\def\iccvPaperID{253} 

\ificcvfinal\fi
\begin{document}

\title{Photographic Image Synthesis with Cascaded Refinement Networks}

\setcounter{footnote}{1}

\author{Qifeng Chen\thanks{Intel Labs}~~\thanks{Stanford University}
\and
Vladlen Koltun\footnotemark[2]
}

\makeatletter
\g@addto@macro\@maketitle{
  \begin{figure}[H]
  \setlength{\linewidth}{\textwidth}
  \setlength{\hsize}{\textwidth}
  \vspace{-7mm}
  \centering
 	\begin{tabular}{@{}c@{\hspace{1mm}}c@{}}
  \includegraphics[width=0.493\textwidth]{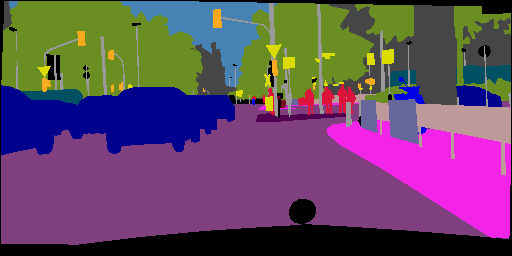} &
  \includegraphics[width=0.493\textwidth]{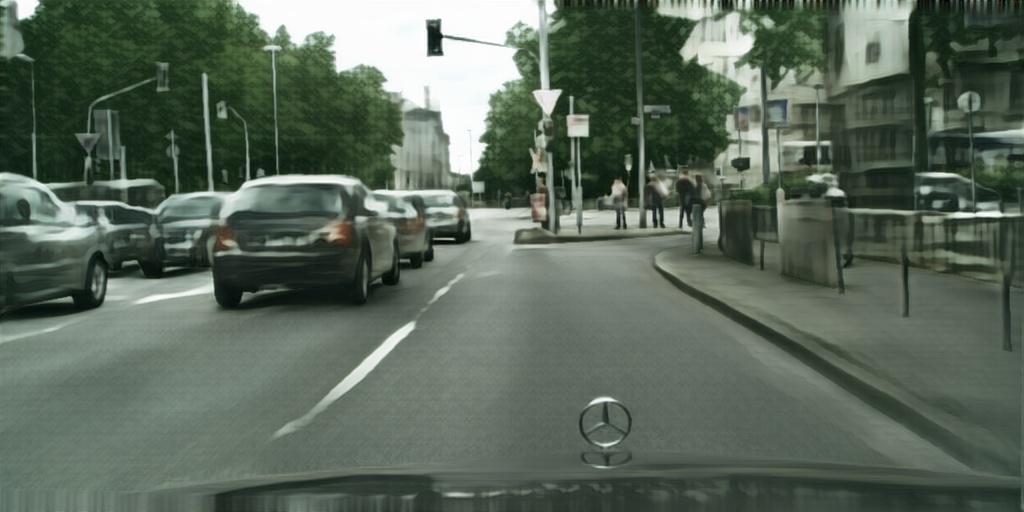}\vspace{0mm}\\
  \includegraphics[width=0.493\textwidth]{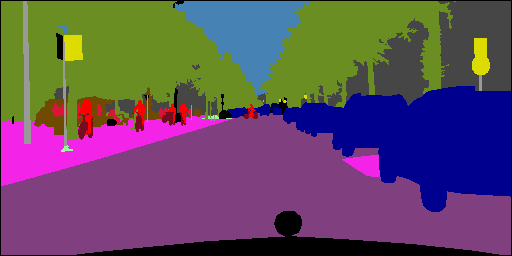} &
  \includegraphics[width=0.493\textwidth]{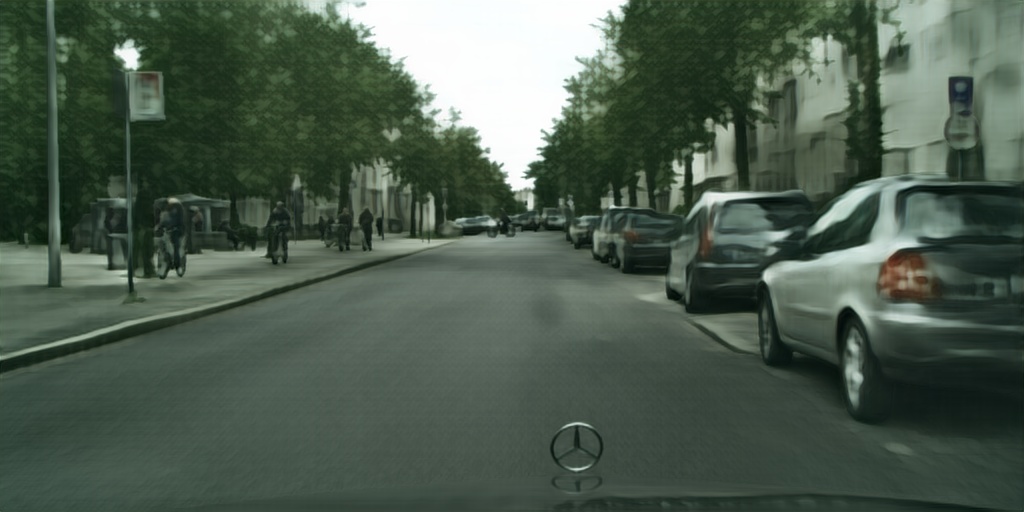}\\
\small (a) Input semantic layouts & \small (b) Synthesized images
 	\end{tabular}
  \vspace{1mm}
  \caption{Given a pixelwise semantic layout, the presented model synthesizes an image that conforms to this layout. (a) Semantic layouts from the Cityscapes dataset of urban scenes; semantic classes are coded by color. (b) Images synthesized by our model for these layouts. The layouts shown here and throughout the paper are from the validation set and depict scenes from new cities that were never seen during training. Best viewed on the screen.}
  \label{fig:teaser}
  \end{figure}
}
\makeatother

\maketitle

\begin{abstract}
We present an approach to synthesizing photographic images conditioned on semantic layouts. Given a semantic label map, our approach produces an image with photographic appearance that conforms to the input layout. The approach thus functions as a rendering engine that takes a two-dimensional semantic specification of the scene and produces a corresponding photographic image. Unlike recent and contemporaneous work, our approach does not rely on adversarial training. We show that photographic images can be synthesized from semantic layouts by a single feedforward network with appropriate structure, trained end-to-end with a direct regression objective. The presented approach scales seamlessly to high resolutions; we demonstrate this by synthesizing photographic images at 2-megapixel resolution, the full resolution of our training data. Extensive perceptual experiments on datasets of outdoor and indoor scenes demonstrate that images synthesized by the presented approach are considerably more realistic than alternative approaches.
\end{abstract}


\section{Introduction}
\label{sec:introduction}
\input{tex/introduction.tex}

\section{Related Work}
\label{sec:related}
\input{tex/related-work.tex}

\section{Method}
\label{sec:method}
\input{tex/method.tex}


\section{Baselines}
\label{sec:baselines}
\input{tex/baselines.tex}

\section{Experiments}
\label{sec:experiments}
\input{tex/experiments.tex}

\section{Conclusion}
\label{sec:discussion}
\input{tex/discussion.tex}

\balance

{\small
\bibliographystyle{ieee}
\bibliography{paper}
}

\end{document}

%% file: vcl-shortcuts.tex
\usepackage{times}
\usepackage{epsfig}
\usepackage{graphicx}
\usepackage{float}
\usepackage{wrapfig}
\usepackage{amsmath,amssymb,amsthm}
\usepackage{algorithm,algorithmicx,algpseudocode}
\usepackage{bm,xspace}
\usepackage{comment}
\usepackage{verbatim}
\usepackage{multirow}
\usepackage{balance}
\usepackage{url}
\usepackage{booktabs}
\usepackage{etoolbox,siunitx}
\usepackage{calc}
\usepackage{pifont,hologo}

\newcommand{\ra}[1]{\renewcommand{\arraystretch}{#1}}

\usepackage[super]{nth}
\usepackage{nicefrac}
\robustify\bfseries

\def\dD{\mathcal{D}}

\def\lL{\mathcal{L}}

\def\Re{\mathbb{R}}

\DeclareMathSymbol{@}{\mathord}{letters}{"3B}

\newcommand\timess{\mathbin{\!\times\!}}

\newcommand\mypara[1]{\vspace{1mm}\noindent\textbf{#1}}

\def\latex/{\LaTeX}
\def\bibtex/{\hologo{BibTeX}}


%% file: tex/introduction.tex
Consider the semantic layouts in Figure~\ref{fig:teaser}. A skilled painter could draw images that depict urban scenes that conform to these layouts. Highly trained craftsmen can even create paintings that approach photorealism~\cite{Letze2013}. Can we train computational models that have this ability? Given a semantic layout of a novel scene, can an artificial system synthesize an image that depicts this scene and looks like a photograph?

This question is connected to central problems in computer graphics and artificial intelligence. First, consider the problem of photorealism in computer graphics. A system that synthesizes photorealistic images from semantic layouts would in effect function as a kind of rendering engine that bypasses the laborious specification of detailed three-dimensional geometry and surface reflectance distributions, and avoids computationally intensive light transport simulation~\cite{Pharr2016}. A direct synthesis approach could not immediately replace modern rendering engines, but would indicate that an alternative route to photorealism may be viable and could some day complement existing computer graphics techniques.

Our second source of motivation is the role of mental imagery and simulation in human cognition~\cite{Kosslyn2006}. Mental imagery is believed to play an important role in planning and decision making. The level of detail and completeness of mental imagery is a matter of debate, but its role in human intelligence suggests that the ability to synthesize photorealistic images may support the development of artificial intelligent systems~\cite{Markman2009}.


In this work, we develop a model for photographic image synthesis from pixelwise semantic layouts. Our model is a convolutional network, trained in a supervised fashion on pairs of photographs and corresponding semantic layouts. Such pairs are provided with semantic segmentation datasets~\cite{Cordts2016}. We use them not to infer semantic layouts from photographs, but to synthesize photographs from semantic layouts. In this sense our problem is the inverse of semantic segmentation. Images synthesized by our model are shown in Figure~\ref{fig:teaser}.

We show that photographic images can be synthesized directly by a single feedforward convolutional network trained to minimize a regression loss. This departs from much recent and contemporaneous work, which uses adversarial training of generator-discriminator dyads~\cite{DosovitskiyBrox2016,Isola2017,Nguyen2016,Salimans2016}. We show that direct supervised training of a single convolutional network can yield photographic images. This bypasses adversarial training, which is known to be ``massively unstable''~\cite{ArjovskyBottou2017}.
Furthermore, the presented approach scales seamlessly to high image resolutions. We synthesize images with resolution up to 2 megapixels \mbox{($1024 \timess 2048$)}, the full resolution of our training data.
Doubling the output resolution and generating appropriate details at that resolution amounts to adding a single module to our end-to-end model.

We conduct careful perceptual experiments using the Amazon Mechanical Turk platform, comparing the presented approach to a range of baselines. These experiments clearly indicate that images synthesized by our model are significantly more realistic than images synthesized by alternative approaches.

%% file: tex/related-work.tex
\begin{figure*}[t]
\centering
\begin{tabular}{@{}c@{\hspace{2mm}}c@{\hspace{2mm}}c@{}}
\rotatebox{90}{\hspace{11mm}Input layout}&
\includegraphics[width=0.46\linewidth]{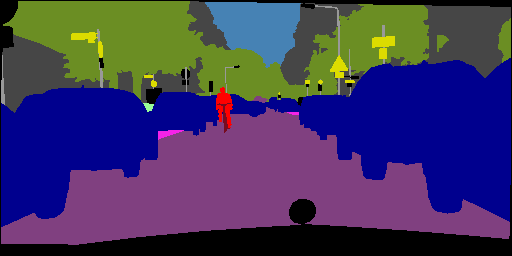} &
\includegraphics[width=0.46\linewidth]{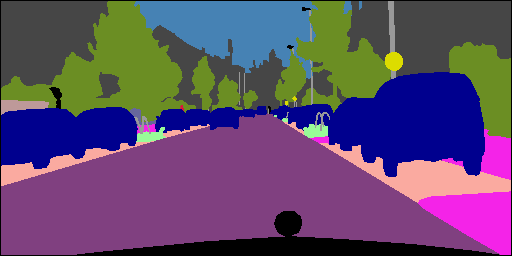}\\
\rotatebox{90}{\hspace{12.5mm}Our result}&
\includegraphics[width=0.46\linewidth]{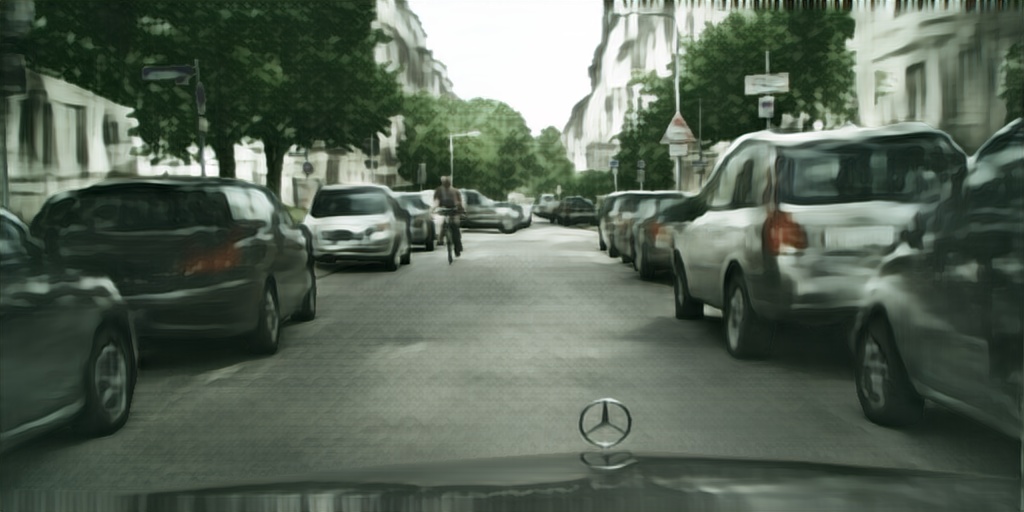}&
\includegraphics[width=0.46\linewidth]{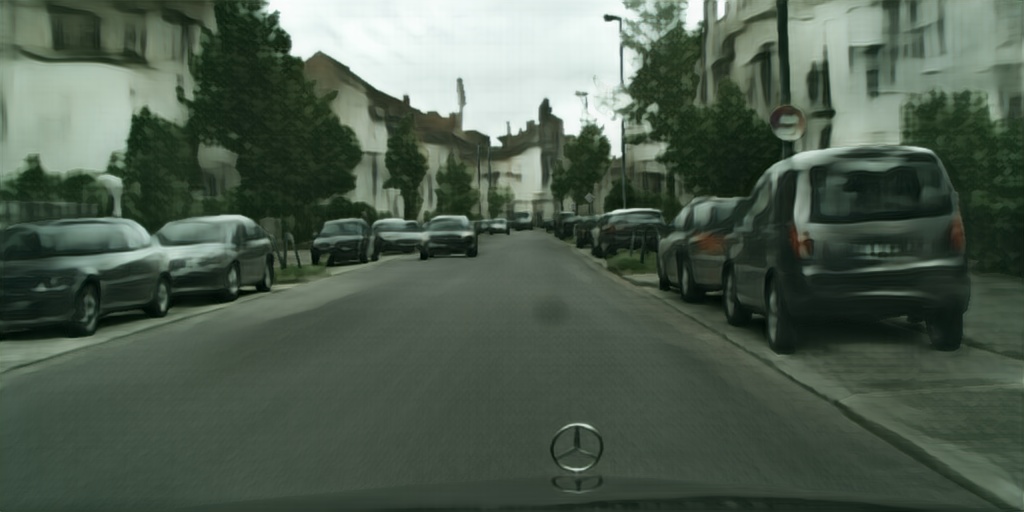} \\
\rotatebox{90}{\hspace{12mm}Isola et al.}&
\includegraphics[width=0.46\linewidth]{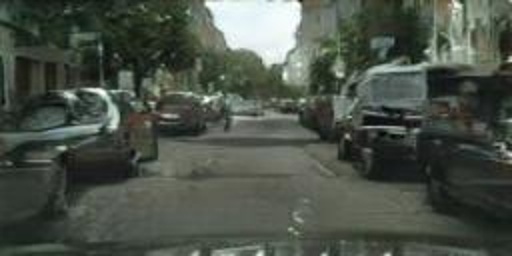}&
\includegraphics[width=0.46\linewidth]{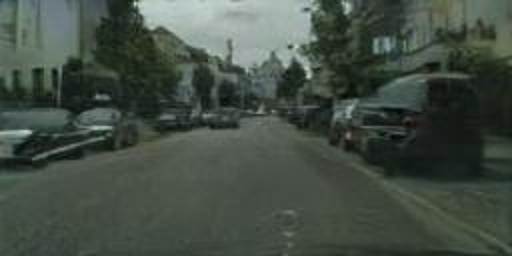}
\end{tabular}
\vspace{0.5mm}
\caption{Comparison to the approach of Isola et al.~\cite{Isola2017}. Zoom in for details.}
\label{fig:comparison}
\vspace{-1mm}
\end{figure*}

The most prominent contemporary approach to image synthesis is based on generative adversarial networks (GANs)~\cite{Goodfellow2014}. In the original work of Goodfellow et al.~\cite{Goodfellow2014}, GANs were used to synthesize MNIST digits and $32\timess 32$ images that aimed to reproduce the appearance of different classes in the CIFAR-10 dataset. Denton et al.~\cite{Denton2015} proposed training multiple separate GANs, one for each level in a Laplacian pyramid. Each model is trained independently to synthesize details at its scale. Assembling separately-trained models in this fashion enabled the authors to synthesize smoother images and to push resolution up to $96\timess 96$.
This work is an important precursor to ours in that multi-scale refinement is a central characteristic of our approach. Key differences are that we train a single model end-to-end to directly synthesize the output image, and that no adversarial training is used.

Radford et al.~\cite{Radford2016} remark that ``Historical attempts to scale up GANs using CNNs to model images have been unsuccessful'' and describe a number of modifications that enable scaling up adversarial training to $64\timess 64$ images. Salimans et al.~\cite{Salimans2016} also tackle the instability of GAN training and describe a number of heuristics that encourage convergence. The authors synthesize $128\timess 128$ images that possess plausible low-level statistics.
Nevertheless, as observed in recent work and widely known in the folklore, GANs ``remain remarkably difficult to train'' and ``approaches to attacking this problem still rely on heuristics that are extremely sensitive to modifications''~\cite{ArjovskyBottou2017}. (See also~\cite{Metz2017}.) Our work demonstrates that these difficulties can be avoided in the setting we consider.

Dosovitskiy et al.~\cite{Dosovitskiy2016} train a ConvNet to generate images of 3D models, given a model ID and viewpoint. The network thus acts directly as a rendering engine for the 3D model. This is also an important precursor to our work as it uses direct feedforward synthesis through a network trained with a regression loss. Our model, loss, and problem setting are different, enabling synthesis of sharper higher-resolution images of scenes without 3D models.

Dosovitskiy and Brox~\cite{DosovitskiyBrox2016} introduced a family of composite loss functions for image synthesis, which combine regression over the activations of a fixed ``perceiver'' network with a GAN loss. Networks trained using these composite loss functions were applied to synthesize preimages that induce desired excitation patterns in image classification models~\cite{DosovitskiyBrox2016} and images that excite specific elements in such models~\cite{Nguyen2016}. In recent work, networks trained using these losses were applied to generate diverse sets of $227\timess 227$ images, to synthesize images for given captions, and to inpaint missing regions~\cite{Nguyen2017}. These works all rely on the aforementioned composite losses, which require balancing the adversarial loss with a regression loss. Our work differs in that GANs are not used, which simplifies the training procedure, architecture, and loss.

Isola et al.~\cite{Isola2017} consider a family of problems that include the image synthesis problem we focus on. The paper of Isola et al.~appeared on arXiv during the course of our research.
It provides an opportunity to compare our approach to a credible alternative that was independently tested on the same data. Like a number of aforementioned formulations, Isola et al.~use a composite loss that combines a GAN and a regression term.
The authors use the Cityscapes dataset and synthesize $256\timess 256$ images for given semantic layouts.
In comparison, our simpler direct formulation yields much more realistic images and scales seamlessly to high resolutions.
A qualitative comparison is shown in Figure~\ref{fig:comparison}.

Reed et al.~\cite{Reed2016:ICML} synthesize $64\timess 64$ images of scenes that are described by given sentences. Mansimov et al.~\cite{Mansimov2015} describe a different model that generates $32\timess 32$ images that aim to fit sentences. Yan et al.~\cite{Yan2016:ECCV} generate $64\timess 64$ images of faces and birds with given attributes. Reed et al.~\cite{Reed2016:NIPS} synthesize $128\timess 128$ images of birds and people conditioned on text descriptions and on spatial constraints such as bounding boxes or keypoints. Wang and Gupta~\cite{WangGupta2016} synthesize $128\timess 128$ images of indoor scenes by factorizing the image generation process into synthesis of a normal map and subsequent synthesis of a corresponding color image. Most of these works use GANs, with the exception of Yan et al.~\cite{Yan2016:ECCV} who use variational autoencoders and Mansimov et al.~\cite{Mansimov2015} who use a recurrent attention-based model~\cite{Gregor2015}. Our problem statement is different in that our input is a pixelwise semantic layout, and our technical approach differs substantially in that a single feedforward convolutional network is trained end-to-end to synthesize a high-resolution image.

A line of work considers synthesis of future frames in video. Srivastava et al.~\cite{Srivastava2015} train a recurrent network for this purpose. Mathieu et al.~\cite{Mathieu2016} build on the work of Denton et al.~\cite{Denton2015} and use a composite loss that combines an adversarial term with regression penalties on colors and gradients. Oh et al.~\cite{Oh2015} predict future frames in Atari games conditioned on the player's action. Finn et al.~\cite{Finn2016} explicitly model pixel motion and also condition on action. Vondrick et al.~\cite{Vondrick2016} learn a model of scene dynamics and use it to synthesize video sequences from single images. Xue et al.~\cite{Xue2016} develop a probabilistic model that enables synthesizing multiple plausible video sequences. In these works, a color image is available as a starting point for synthesis. Video synthesis can be accomplished by advecting the content of this initial image.
In our setting, photographic scene appearance must be synthesized without such initialization.

Researchers have also studied image inpainting~\cite{Pathak2016}, superresolution~\cite{Bruna2016,Johnson2016,Ledig2016}, novel view synthesis~\cite{Flynn2016,Tatarchenko2016,Zhou2016}, and interactive image manipulation~\cite{Zhu2016}. In these problems, photographic content is given as input, whereas we are concerned with synthesizing photographic images from semantic layouts alone.

%% file: tex/method.tex
\subsection{Preliminaries}

Consider a semantic layout $L \in \{0,1\}^{m\times n \times c}$, where $m\timess n$ is the pixel resolution and $c$ is the number of semantic classes. Each pixel in $L$ is represented by a one-hot vector that indicates its semantic label: ${L(i,j) \in \{0,1\}^c}$ s.t. ${\sum_{p}L(i,j,p)=1}$. One of the $c$ possible labels is `void', which indicates that the semantic class of the pixel is not specified.
Our goal is to train a parametric mapping $g$ that given a semantic layout $L$ produces a color image ${I \in \Re^{m\times n \times 3}}$ that conforms to $L$.

In the course of this project we have experimented with a large number of network architectures.
As a result of these experiments, we have identified three characteristics that are important for synthesizing photorealistic images. We review these characteristics before describing our solution.

\mypara{Global coordination.}
Globally consistent structure is essential for photorealism. Many objects exhibit nonlocal structural relationships, such as symmetry. For example, if the network synthesizes a red light on the left side of a car, then the corresponding light on the right should also be red. This distinguishes photorealistic image synthesis from texture synthesis, which can leverage statistical stationarity~\cite{PortillaSimoncelli2000}. Our model is based on multi-resolution refinement. The synthesis begins at extremely low resolution ($4\timess 8$ in our implementation). Feature maps are then progressively refined. Thus global structure can be coordinated at lower octaves, where even distant object parts are represented in nearby feature columns. These decisions are then refined at higher octaves.


\mypara{High resolution.}
To produce truly photorealistic results, a model must be able to synthesize high-resolution images. Low resolution is akin to myopic vision in that fine visual features are not discernable. The drive to high image and video resolutions in multiple industries is a testament to resolution's importance. Our model synthesizes images by progressive refinement, and going up an octave in resolution (e.g., from 512p to 1024p) amounts to adding a single refinement module. The entire cascade of refinement modules is trained end-to-end.

\mypara{Memory.}
We conjecture that high model capacity is essential for synthesizing high-resolution photorealistic images. Human hyperrealistic painters use photographic references as external memory of detailed object appearance~\cite{Letze2013}. The best existing image compression techniques require millions of bits of information to represent the content of a single high-resolution image: there exists no known way to reconstruct a given photograph at high fidelity from a lower-capacity representation~\cite{Sayood2012}. In order for our model to be able to synthesize diverse scenes from a given domain given only semantic layouts as input, the capacity of the model must be sufficiently high to be able to reproduce the detailed photographic appearance of many objects. We expect a successful model to reproduce images in the training set extremely well (memorization) and also to apply the learned representations to novel layouts (generalization). This requires high model capacity. Our design is modular and the capacity of the model can be expanded as allowed by hardware. The network used in most of our experiments has 105M parameters and maximizes available GPU memory. We have consistently found that increasing model capacity increases image quality.

\begin{figure}[t]
  \centering
  \includegraphics[width=1\linewidth]{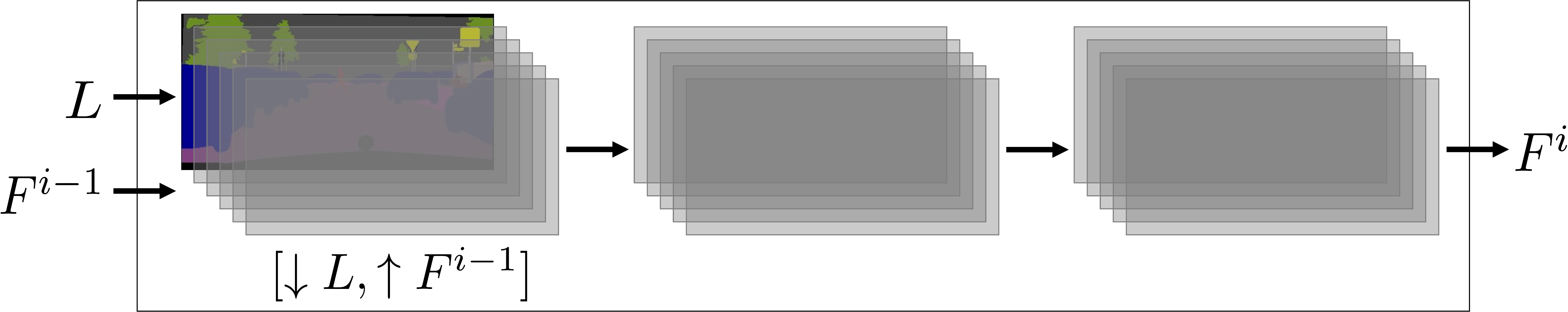}
  \caption{A single refinement module.}
\label{fig:module}
\vspace{-1mm}
\end{figure}

\subsection{Architecture}
\label{sec:architecture}

The Cascaded Refinement Network (CRN) is a cascade of refinement modules. Each module $M^i$ operates at a given resolution. In our implementation, the resolution of the first module ($M^0$) is $4\timess 8$. Resolution is doubled between consecutive modules (from $M^{i-1}$ to $M^i$). Let $w_i \timess h_i$ be the resolution of module $i$.

The first module, $M^0$, receives the semantic layout $L$ as input (downsampled to $w_0 \timess h_0$) and produces a feature layer $F^0$ at resolution $w_0 \timess h_0$ as output. All other modules $M^i$ (for ${i\ne 0}$) are structurally identical: $M^i$ receives a concatenation of the layout $L$ (downsampled to $w_i \timess h_i$) and the feature layer $F^{i-1}$ (upsampled to $w_i \timess h_i$) as input, and produces feature layer $F^i$ as output. We denote the number of feature maps in $F^i$ by $d_i$.

Each module $M^i$ consists of three feature layers: the input layer, an intermediate layer, and the output layer. This is illustrated in Figure~\ref{fig:module}. The input layer has dimensionality ${w_i \timess h_i \timess (d_{i-1}+c)}$ and is a concatenation of the downsampled semantic layout $L$ ($c$ channels) and a bilinearly upsampled feature layer $F^{i-1}$ ($d_{i-1}$ channels). Note that we do not use upconvolutions because upconvolutions tend to introduce characteristic artifacts~\cite{Odena2016}.
The intermediate layer and the output layer both have dimensionality $w_i \timess h_i \timess d_i$. Each layer is followed by $3\timess 3$ convolutions, layer normalization~\cite{Ba2016}, and LReLU nonlinearity~\cite{Maas2013}.

The output layer $F^{\bar\imath}$ of the final module $M^{\bar\imath}$ is not followed by normalization or nonlinearity. Instead, a linear projection ($1\timess 1$ convolution) is applied to map $F^{\bar\imath}$ (dimensionality ${w_{\bar\imath} \timess h_{\bar\imath} \timess d_{\bar\imath}}$) to the output color image (dimensionality ${w_{\bar\imath} \timess h_{\bar\imath} \timess 3}$).
The total number of refinement modules in a cascade depends on the output resolution. For our main experiments on the high-resolution Cityscapes dataset, the number of modules is 9, accounting for a resolution increase from $4\timess 8$ to $1024 \timess 2048$.
For the number of feature maps $d_i$, we use 1024 for $i = 0..4$, 512 for $i=5,6$, 128 for $i=7$, and 32 for $i=8$.

\subsection{Training}

The CRN is trained in a supervised fashion on a semantic segmentation dataset $\dD = \{(I,L)\}$. A semantic layout $L$ is used as input and the corresponding color image $I$ as output. This can be thought of as ``inverse semantic segmentation''. It is an underconstrained one-to-many inverse problem. We will generally refer to $I$ as a ``reference image'' rather than ``ground truth'', since many valid photographic images could have yielded the same semantic layout.


Given the underconstrained nature of the problem, using an appropriate loss function is critical, as observed in prior work on image synthesis. Simply comparing the pixel colors of the synthesized image and the reference image could severely penalize perfectly realistic outputs. For example, synthesizing a white car instead of a black car would induce a very high loss. Instead we adopt the ``content representation'' of Gatys et al.~\cite{Gatys2016}, also referred to as a perceptual loss or feature matching~\cite{Bruna2016,DosovitskiyBrox2016,Johnson2016,Ledig2016,Nguyen2016,Nguyen2017}. The basic idea is to match activations in a visual perception network that is applied to the synthesized image and separately to the reference image.

Let $\Phi$ be a trained visual perception network (we use VGG-19~\cite{SimonyanZisserman2015}). Layers in the network represent an image at increasing levels of abstraction: from edges and colors to objects and categories.
Matching both lower-layer and higher-layer activations in the perception network guides the synthesis network to learn both fine-grained details and more global part arrangement.

Let $\{\Phi_l\}$ be a collection of layers in the network $\Phi$, such that $\Phi_0$ denotes the input image. Each layer is a three-dimensional tensor.
For a training pair ${(I,L) \in \dD}$, our loss is
\begin{equation}
\lL_{I,L}(\theta) = \sum_l{\lambda_l\| \Phi_l(I)-\Phi_l(g(L;\theta))\|_1}.
\label{eq:loss}
\end{equation}
Here $g$ is the image synthesis network being trained and $\theta$ is the set of parameters of this network. The hyperparameters $\{\lambda_l\}$ balance the contribution of each layer $l$ to the loss.



For layers $\Phi_l$ ($l\! \ge\! 1$) we use `conv1\_2', `conv2\_2', `conv3\_2', `conv4\_2', and `conv5\_2' in VGG-19~\cite{SimonyanZisserman2015}. The hyperparameters $\{\lambda_l\}$ are set automatically. They are initialized to the inverse of the number of elements in each layer. After 100 epochs, $\{\lambda_l\}$ are rescaled to normalize the expected contribution of each term ${\| \Phi_l(I)-\Phi_l(g(L;\theta))\|_1}$ to the loss.



\subsection{Synthesizing a diverse collection}
\label{sec:diversity}

The architecture and training procedure described so far synthesize a single image for a given input $L$. In our experiments this already yields good results.
However, since a given semantic layout can correspond to many images, it also makes sense to generate a diverse set of images as output.
Conditional synthesis of diverse images can be approached as a stochastic process~\cite{Nguyen2017}. We take a different tack and modify the network to emit a collection of images in one shot, with a modified loss that encourages diversity within the collection.


Specifically, we change the number of output channels from $3$ to $3k$, where $k$ is the desired number of images. Each consecutive 3-tuple of channels forms an image. Now consider the loss. If loss (\ref{eq:loss}) is applied independently to each output image, the $k$ synthesized images will be identical. Our first modification is to consider the set of $k$ outputs together and define the loss of the whole collection in terms of the \emph{best} synthesized image. Let $g_u(L;\theta)$ be the $u^{\text{th}}$ image in the synthesized collection. Our first version of the modified loss is based on the hindsight loss developed for multiple choice learning~\cite{Guzman-Rivera2012}:
\begin{equation}
\min_u{\sum_l{\lambda_l\| \Phi_l(I)-\Phi_l(g_u(L;\theta))\|_1}}.
\label{eq:loss-diverse1}
\end{equation}
By considering only the best synthesized image, this loss encourages the network to spread its bets and cover the space of images that conform to the input semantic layout. The loss is structurally akin to the $k$-means clustering objective, which only considers the closest centroid to each datapoint and thus encourages the centroids to spread and cover the dataset.


We further build on this idea and formulate a loss that considers a virtual collection of up to $k^c$ images. (Recall that $c$ is the number of semantic classes.) Specifically, for each semantic class $p$, let $L_p$ denote the corresponding channel $L(\cdot,\cdot,p)$ in the input label map. We now define a more powerful diversity loss as
\begin{equation}
\resizebox{.9\hsize}{!}{$\displaystyle{
\sum_{p=1}^c{\min_u{\sum_l{\lambda_l\sum_j \big\|L_p^l \odot \big(\Phi_l^j(I)-\Phi_l^j(g_u(L;\theta))\big)\big\|_1}}},
}$}
\label{eq:loss-diverse2}
\end{equation}
where $\Phi_l^j$ is the $j^{\text{th}}$ feature map in $\Phi_l$, $L_p^l$ is the mask $L_p$ downsampled to match the resolution of $\Phi_l$, and $\odot$ is the Hadamard product. This loss in effect constructs a virtual image by adaptively taking the best synthesized content for each semantic class from the whole collection, and scoring the collection based on this assembled image.

%% file: tex/baselines.tex
The approach presented in Section~\ref{sec:method} is far from the first we tried. In this section we describe a number of alternative approaches that will be used as baselines in Section~\ref{sec:experiments}.

\mypara{GAN and semantic segmentation.}
Our first baseline is consistent with current trends in the research community. It combines a GAN with a semantic segmentation objective. The generator is trained to synthesize an image that fools the discriminator~\cite{Goodfellow2014}. An additional term in the loss specifies that when the synthesized image is given as input to a pretrained semantic segmentation network, it should produce a label map that is as close to the input layout $L$ as possible. The GAN setup follows the work of Radford et al.~\cite{Radford2016}. The input to the generator is the semantic layout $L$. For the semantic segmentation network, we use publicly available networks that were pretrained for the Cityscapes dataset~\cite{YuKoltun2016} and the NYU dataset~\cite{Long2015}. The training objective combines the GAN loss and the semantic segmentation (pixelwise cross-entropy) loss.


\mypara{Full-resolution network.}
Our second baseline is a feedforward convolutional network that operates at full resolution. This baseline uses the same loss as the CRN described in Section~\ref{sec:method}. The only difference is the network architecture. In particular, we have experimented with variants of the multi-scale context aggregation network~\cite{YuKoltun2016}. An appealing property of this network is that it retains high resolution in the intermediate layers, which we hypothesized to be helpful for photorealistic image synthesis. The original architecture described in~\cite{YuKoltun2016} did not yield good results and is not well-suited to our problem, because the input semantic layouts are piecewise constant and the network of~\cite{YuKoltun2016} begins with a small receptive field. We obtained much better results with the inverse architecture: start with large dilation and decrease it by a factor of 2 in each layer. This can be viewed as a full-resolution counterpart to the CRN, based on dilating the filters instead of scaling the feature maps. One of the drawbacks of this approach is that all intermediate feature layers are at full image resolution and have a high memory footprint. Thus the ratio of capacity (number of parameters) to memory footprint is much lower than in the CRN. This high memory footprint of intermediate layers also constrains the resolution to which this approach can scale: with 10 layers and 256 feature maps per layer, the maximal resolution that could be trained with available GPU memory is $256\timess 512$.


\mypara{Encoder-decoder.}
Our third baseline is an encoder-decoder network, the u-net~\cite{Ronneberger2015}. This network is also trained with the same loss as the CRN. It is thus an additional baseline that evaluates the effect of using the CRN versus a different architecture, when everything else (loss, training procedure) is held fixed.

\mypara{Image-space loss.}
Our next baseline controls for the feature matching loss used to train the CRN. Here we use exactly the same architecture as in Section~\ref{sec:method}, but use only the first layer $\Phi_0$ (image color) in the loss:
\begin{equation}
\lL_{I,L}(\theta) = \sum_l{\lambda_l\| I-g(L;\theta)\|_1}.
\label{eq:loss-color}
\end{equation}


\mypara{Image-to-image translation.}
Our last baseline is the contemporaneous approach of Isola et al., the implementation and results of which are publicly available~\cite{Isola2017}. This approach uses a conditional GAN and is representative of the dominant stream of research in image synthesis. The generator is an encoder-decoder~\cite{Ronneberger2015}. The GAN setup is derived from the work of Radford et al.~\cite{Radford2016}.

%% file: tex/experiments.tex
\begin{table*}
   \centering
  \setlength{\tabcolsep}{4mm}
   \ra{1.1}
\resizebox{1\linewidth}{!}{
  \begin{tabular}{@{}l@{\hspace{8mm}}ccccc@{}}
     \toprule
    &  Image-space loss & GAN+SemSeg & Isola et al.~\cite{Isola2017} & Encoder-decoder  & Full-resolution network\\       \midrule
   Cityscapes &     99.7\%   	&   98.5\%   &  96.9\% &  78.3\% &67.7\%  \\
   NYU &  91.4\% & 82.3\% & 77.2\% &  71.2\% &65.8\%\\
 \bottomrule
\end{tabular}
}
\vspace{0.5mm}
\caption{Results of pairwise comparisons of images synthesized by models trained on the Cityscapes and NYU datasets. Each column compares our approach with one of the baselines. Each cell lists the fraction of pairwise comparisons in which images synthesized by our approach were rated more realistic than images synthesized by the corresponding baseline. Chance is at 50\%.}
\label{table:cityscapes}
\vspace{-1mm}
\end{table*}


\subsection{Experimental procedure}

\mypara{Methodology.}
The most reliable known methodology for evaluating the realism of synthesized images is perceptual experiments with human observers. Such experiments yield quantitative results and have been used in related work~\cite{Denton2015,Ledig2016,Salimans2016}. There have also been attempts to design automatic measures that evaluate realism without humans in the loop. For example, Salimans et al.\ ran a pretrained image classification network on synthesized images and analyzed its predictions~\cite{Salimans2016}. We experimented with such automatic measures (for example using pretrained semantic segmentation networks) and found that they can all be fooled by augmenting any baseline to also optimize for the evaluated measure; the resulting images are not more realistic but score very highly~\cite{Goodfellow2015,Nguyen2015}. Well-designed perceptual experiments with human observers are more reliable. We therefore use carefully designed perceptual experiments for quantitative evaluation. We will release our complete implementation and experimental setup so that our experiments can be replicated by others.

All experiments use pairwise A/B tests deployed on the Amazon Mechanical Turk (MTurk) platform. Similar protocols have been used to evaluate the realism of 3D reconstructions~\cite{Choi2015,Shan2013}.
Each MTurk job involves a batch of roughly 100 pairwise comparisons, along with sentinel pairs that test whether the worker is attentive and diligent. Each pair contains two images synthesized for the same label map by two different approaches (or a corresponding reference image from the dataset). The workers are asked to select the more realistic image in each pair. The images are all shown at the same resolution ($200\timess 400$). The comparisons are randomized across conditions and both the left-right order and the order within a job are randomized.

Two types of experiments are conducted. In the first, images are shown for unlimited time and the worker is free to spend as much time as desired on each pair. In the second, each pair is shown for a randomly chosen duration between $\frac{1}{8}$ and $8$ seconds. This evaluates how quickly the relative realism of different image pairs can be established.

The experimental setup is further detailed in the supplement and is demonstrated in supplementary videos.

\mypara{Datasets.}
We use two datasets with pixelwise semantic labels, one depicting outdoor scenes and one depicting indoor scenes. Our primary dataset is Cityscapes, which has become the dominant semantic segmentation dataset due to the quality of the data~\cite{Cordts2016}. We train on the training set (3K images) and evaluate on the validation set (500 images). (Evaluating ``inverse semantic segmentation'' on the test set is impossible because the label maps are not provided.)
Our second dataset is the older NYU dataset of indoor scenes~\cite{Silberman2012}. This dataset is smaller and the images are VGA resolution.
Note that we do not use the depth data in the NYU dataset, only the semantic layouts and the color images. We use the first 1200 of the 1449 labeled images for training and the remaining 249 for testing.



\subsection{Results}

\begin{figure*}[t]
\centering
\begin{tabular}{@{}c@{\hspace{0.8mm}}c@{\hspace{0.8mm}}c@{}}
\includegraphics[width=0.33\linewidth]{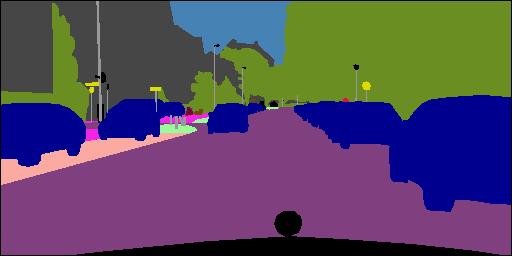}&
\includegraphics[width=0.33\linewidth]{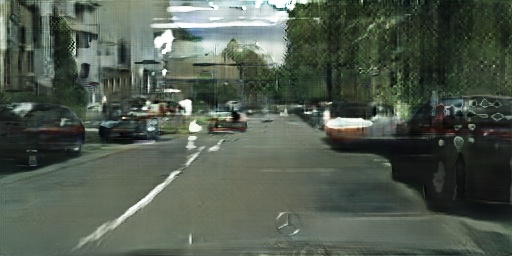}&
\includegraphics[width=0.33\textwidth]{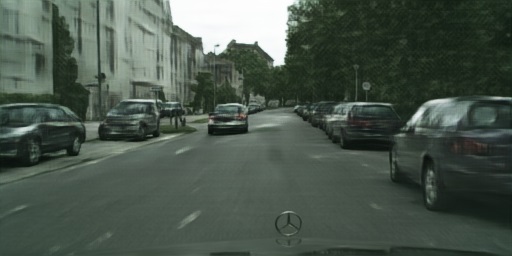}\vspace{-1mm}\\
\small Semantic layout & \small GAN+semantic segmenation & \small Full-resolution network  \vspace{1mm}\\
\includegraphics[width=0.33\linewidth]{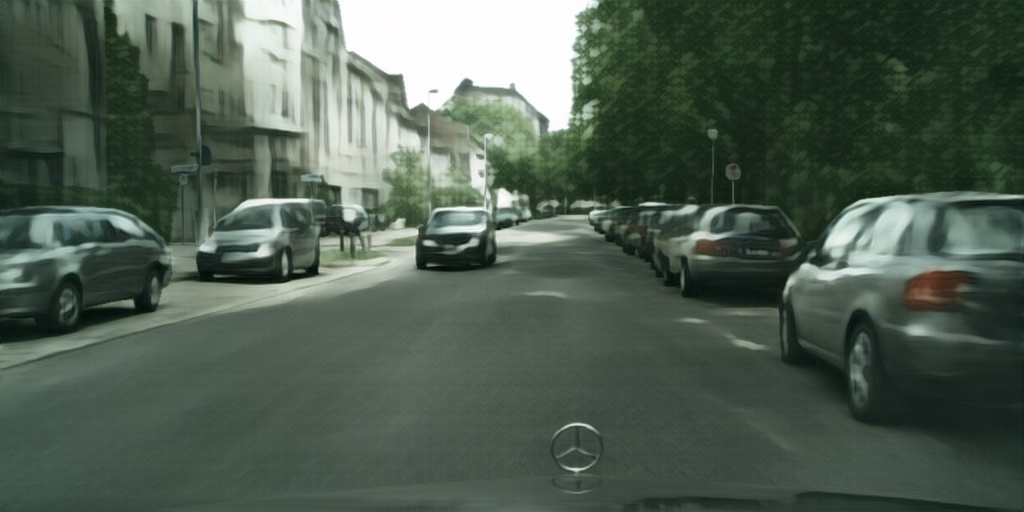}&
\includegraphics[width=0.33\textwidth]{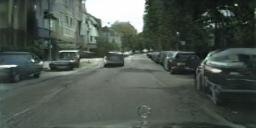}&
\includegraphics[width=0.33\linewidth]{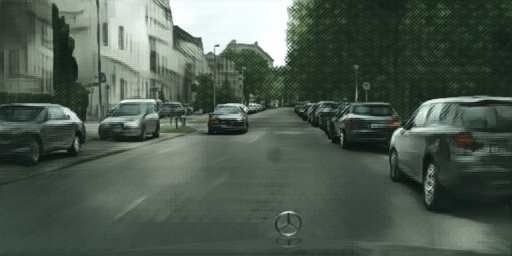}\vspace{-1mm}\\
\small Our result & \small Isola et al.~\cite{Isola2017} & \small Encoder-decoder \vspace{1mm}\\
\end{tabular}
\caption{Qualitative comparison on the Cityscapes dataset.}
\label{fig:visual_cityscapes}
\vspace{3mm}

\centering
\begin{tabular}{@{}c@{\hspace{0.3mm}}c@{\hspace{0.3mm}}c@{\hspace{0.3mm}}c@{\hspace{0.3mm}}c@{}}
\includegraphics[width=0.198\linewidth]{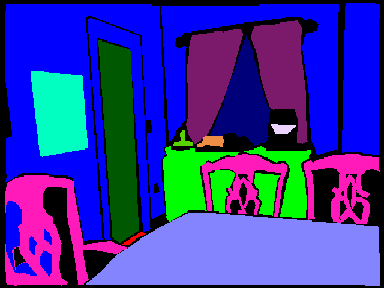}&
\includegraphics[width=0.198\linewidth]{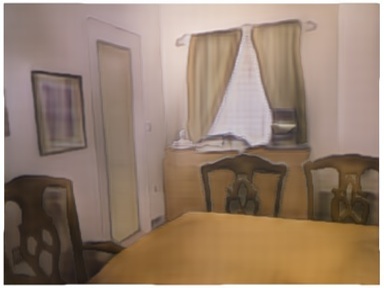}&
\includegraphics[width=0.198\linewidth]{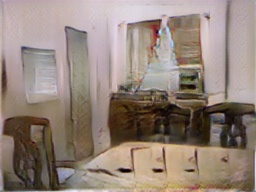}&
\includegraphics[width=0.198\linewidth]{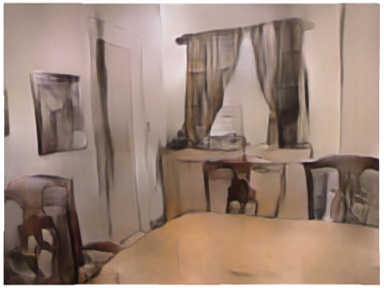}&
\includegraphics[width=0.198\linewidth]{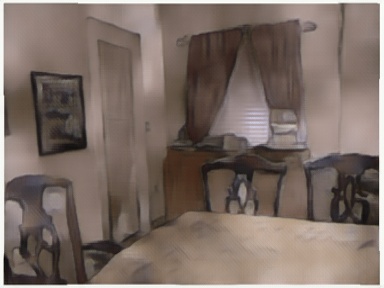}\\
\includegraphics[width=0.198\linewidth]{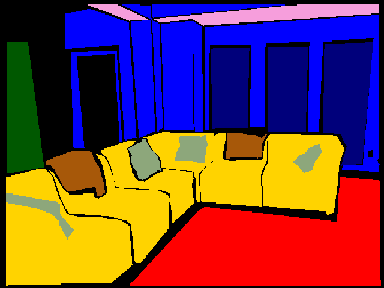}&
\includegraphics[width=0.198\linewidth]{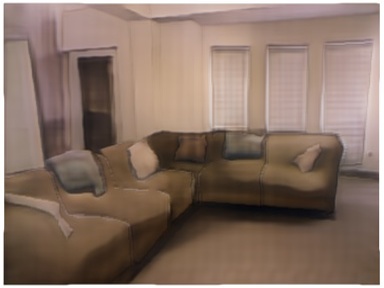}&
\includegraphics[width=0.198\linewidth]{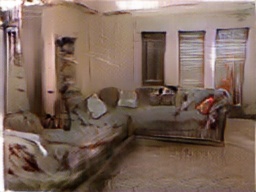}&
\includegraphics[width=0.198\linewidth]{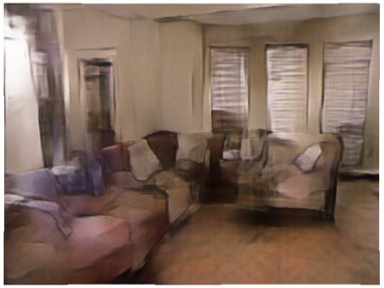}&
\includegraphics[width=0.198\linewidth]{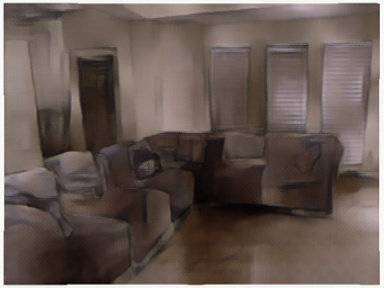}\\
\small Semantic layout & \small Our result & \small Isola et al.~\cite{Isola2017} & \small Full-resolution network & \small Encoder-decoder \vspace{1mm}\\
\end{tabular}
\caption{Qualitative comparison on the NYU dataset.}
\label{fig:visual_nyu}
\vspace{3mm}

\centering
\begin{tabular}{@{}c@{\hspace{0mm}}c@{\hspace{0mm}}c@{\hspace{0mm}}c@{}}
\includegraphics[height=3.25cm]{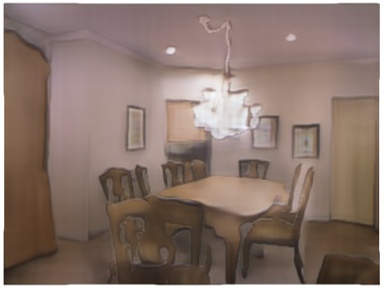}&
\includegraphics[height=3.25cm]{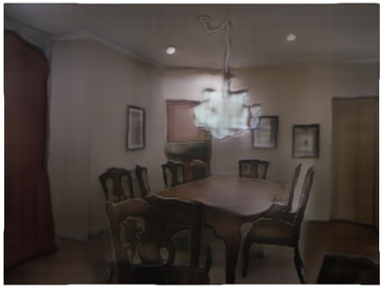}&
\includegraphics[height=3.25cm]{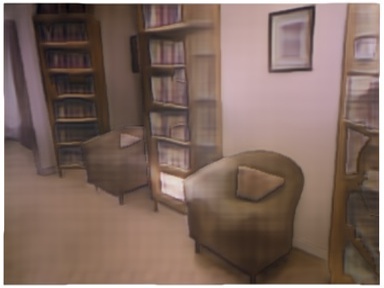}&
\includegraphics[height=3.25cm]{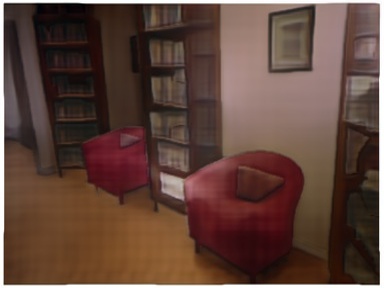}\\
\end{tabular}
\caption{Synthesizing a diverse collection, illustrated on the NYU dataset. Each pair shows two images from a collection synthesized for a given semantic layout.
}
\label{fig:diversity}
\vspace{-2mm}
\end{figure*}

\mypara{Primary experiments.}
Table~\ref{table:cityscapes} reports the results of randomized pairwise comparisons of images synthesized by models trained on the Cityscapes dataset. Images synthesized by the presented approach were rated more realistic than images synthesized by the four alternative approaches. Note that the `image-space loss' baseline uses the same architecture as the CRN and controls for the loss, while the `full-resolution network' and the `encoder-decoder' use the same loss as the CRN and control for the architecture. All results are statistically significant with $p < 10^{-3}$. Compared to the approach of Isola et al.~\cite{Isola2017}, images synthesized by the CRN were rated more realistic in 97\% of the comparisons. Qualitative results are shown in Figure~\ref{fig:visual_cityscapes}.

\begin{figure}[h]
\centering
\includegraphics[width=0.95\linewidth]{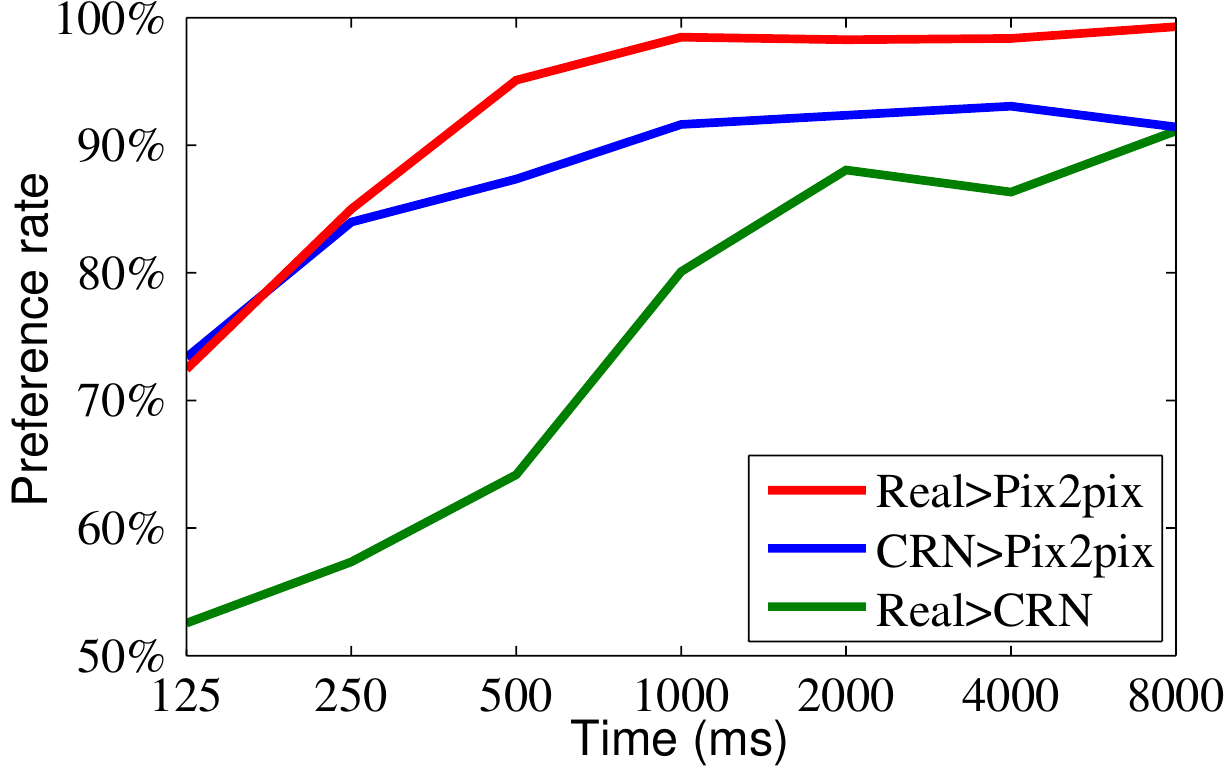}
\caption{Time-limited pairwise comparisons on Cityscapes.}
\label{fig:cityscapes-timed}
\vspace{-1mm}
\end{figure}

Figure~\ref{fig:cityscapes-timed} reports the results of time-limited pairwise comparisons of real Cityscapes images, images synthesized by the CRN, and images synthesized by the approach of Isola et al.~\cite{Isola2017} (referred to as `Pix2pix' following the public implementation). After just $\frac{1}{8}$ of a second, the Pix2pix images are clearly rated less realistic than the real Cityscapes images or the CRN images (72.5\% Real$>$Pix2pix, 73.4\% CRN$>$Pix2pix). On the other hand, the CRN images are on par with real images at that time, as seen both in the Real$>$CRN rate (52.6\%) and in the nearly identical Real$>$Pix2pix and CRN$>$Pix2pix rates.

At 250 milliseconds ($\frac{1}{4}$ of a second), the Real$>$Pix2pix rate rises to 85.0\% while the Real$>$CRN rate is at 57.4\%. The CRN$>$Pix2pix rate is 84.0\%, still nearly identical to Real$>$Pix2pix. At 500 milliseconds, the Real$>$Pix2pix and CRN$>$Pix2pix rates finally diverge, although both are extremely high (95.1\% and 87.4\%, respectively), and the Real$>$CRN rate rises to 64.2\%. Over time, the CRN$>$Pix2pix rate rises above 90\% and the Real$>$Pix2pix rate remains consistently higher than the Real$>$CRN rate.

\mypara{NYU dataset.}
We conduct supporting experiments on the NYU dataset. This dataset is smaller and lower-resolution, so the quality of images synthesized by all approaches is lower. Nevertheless, the differences are still clear. Table~\ref{table:cityscapes} reports the results of randomized pairwise comparisons of images synthesized for this dataset. Images synthesized by the presented approach were again rated consistently more realistic than the baselines. All results are statistically significant with $p < 10^{-3}$. Qualitative results are shown in Figure~\ref{fig:visual_nyu}.

\mypara{Diversity loss.}
For all preceding experiments we have used the feature matching loss specified in Equation~\ref{eq:loss}. The models produced a single image as output, and this image was evaluated against baselines. We now qualitatively demonstrate the effect of the diversity loss described in Section~\ref{sec:diversity}. To this end we trained models that produce image collections as output (9 images at a time). Figure~\ref{fig:diversity} shows pairs of images sampled from the synthesized collections, for different input layouts in the NYU validation set. The figure illustrates that the diversity loss does lead the output channels to spread out and produce different appearances.

%% file: tex/discussion.tex
We have presented a direct approach to photographic image synthesis conditioned on pixelwise semantic layouts. Images are synthesized by a convolutional network trained end-to-end with a regression loss. This direct approach is considerably simpler than contemporaneous work, and produces much more realistic results. We hope that the simplicity of the presented approach can support follow-up work that will further advance realism and explore the applications of photographic image synthesis. Our results, while significantly more realistic than the prior state of the art, are clearly not indistinguishable from real HD images. Exciting work remains to be done to achieve perfect photorealism. If such level of realism is ever achieved, which we believe to be possible, alternative routes for image synthesis in computer graphics will open up.

%% file: paper.bbl
\begin{thebibliography}{10}\itemsep=-1pt

\bibitem{ArjovskyBottou2017}
M.~Arjovsky and L.~Bottou.
\newblock Towards principled methods for training generative adversarial
  networks.
\newblock In {\em ICLR}, 2017.

\bibitem{Ba2016}
J.~L. Ba, J.~R. Kiros, and G.~E. Hinton.
\newblock Layer normalization.
\newblock {\em arXiv:1607.06450}, 2016.

\bibitem{Bruna2016}
J.~Bruna, P.~Sprechmann, and Y.~LeCun.
\newblock Super-resolution with deep convolutional sufficient statistics.
\newblock In {\em ICLR}, 2016.

\bibitem{Choi2015}
S.~Choi, Q.~Zhou, and V.~Koltun.
\newblock Robust reconstruction of indoor scenes.
\newblock In {\em CVPR}, 2015.

\bibitem{Cordts2016}
M.~Cordts, M.~Omran, S.~Ramos, T.~Rehfeld, M.~Enzweiler, R.~Benenson,
  U.~Franke, S.~Roth, and B.~Schiele.
\newblock The {Cityscapes} dataset for semantic urban scene understanding.
\newblock In {\em CVPR}, 2016.

\bibitem{Denton2015}
E.~L. Denton, S.~Chintala, A.~Szlam, and R.~Fergus.
\newblock Deep generative image models using a {Laplacian} pyramid of
  adversarial networks.
\newblock In {\em NIPS}, 2015.

\bibitem{DosovitskiyBrox2016}
A.~Dosovitskiy and T.~Brox.
\newblock Generating images with perceptual similarity metrics based on deep
  networks.
\newblock In {\em NIPS}, 2016.

\bibitem{Dosovitskiy2016}
A.~Dosovitskiy, J.~T. Springenberg, M.~Tatarchenko, and T.~Brox.
\newblock Learning to generate chairs, tables and cars with convolutional
  networks.
\newblock {\em PAMI}, 2016.

\bibitem{Finn2016}
C.~Finn, I.~J. Goodfellow, and S.~Levine.
\newblock Unsupervised learning for physical interaction through video
  prediction.
\newblock In {\em NIPS}, 2016.

\bibitem{Flynn2016}
J.~Flynn, I.~Neulander, J.~Philbin, and N.~Snavely.
\newblock Deep stereo: Learning to predict new views from the world's imagery.
\newblock In {\em CVPR}, 2016.

\bibitem{Gatys2016}
L.~A. Gatys, A.~S. Ecker, and M.~Bethge.
\newblock Image style transfer using convolutional neural networks.
\newblock In {\em CVPR}, 2016.

\bibitem{Goodfellow2014}
I.~J. Goodfellow, J.~Pouget{-}Abadie, M.~Mirza, B.~Xu, D.~Warde{-}Farley,
  S.~Ozair, A.~C. Courville, and Y.~Bengio.
\newblock Generative adversarial nets.
\newblock In {\em NIPS}, 2014.

\bibitem{Goodfellow2015}
I.~J. Goodfellow, J.~Shlens, and C.~Szegedy.
\newblock Explaining and harnessing adversarial examples.
\newblock In {\em ICLR}, 2015.

\bibitem{Gregor2015}
K.~Gregor, I.~Danihelka, A.~Graves, D.~J. Rezende, and D.~Wierstra.
\newblock {DRAW}: A recurrent neural network for image generation.
\newblock In {\em ICML}, 2015.

\bibitem{Guzman-Rivera2012}
A.~Guzm{\'{a}}n{-}Rivera, D.~Batra, and P.~Kohli.
\newblock Multiple choice learning: Learning to produce multiple structured
  outputs.
\newblock In {\em NIPS}, 2012.

\bibitem{Isola2017}
P.~Isola, J.~Zhu, T.~Zhou, and A.~A. Efros.
\newblock Image-to-image translation with conditional adversarial networks.
\newblock In {\em CVPR}, 2017.

\bibitem{Johnson2016}
J.~Johnson, A.~Alahi, and L.~Fei{-}Fei.
\newblock Perceptual losses for real-time style transfer and super-resolution.
\newblock In {\em ECCV}, 2016.

\bibitem{Kosslyn2006}
S.~M. Kosslyn, W.~L. Thompson, and G.~Ganis.
\newblock {\em The Case for Mental Imagery}.
\newblock Oxford University Press, 2006.

\bibitem{Ledig2016}
C.~Ledig, L.~Theis, F.~Huszar, J.~Caballero, A.~P. Aitken, A.~Tejani, J.~Totz,
  Z.~Wang, and W.~Shi.
\newblock Photo-realistic single image super-resolution using a generative
  adversarial network.
\newblock In {\em CVPR}, 2017.

\bibitem{Letze2013}
O.~Letze.
\newblock {\em Photorealism: 50 Years of Hyperrealistic Painting}.
\newblock Hatje Cantz, 2013.

\bibitem{Long2015}
J.~Long, E.~Shelhamer, and T.~Darrell.
\newblock Fully convolutional networks for semantic segmentation.
\newblock In {\em CVPR}, 2015.

\bibitem{Maas2013}
A.~L. Maas, A.~Y. Hannun, and A.~Y. Ng.
\newblock Rectifier nonlinearities improve neural network acoustic models.
\newblock In {\em ICML}, 2013.

\bibitem{Mansimov2015}
E.~Mansimov, E.~Parisotto, L.~J. Ba, and R.~Salakhutdinov.
\newblock Generating images from captions with attention.
\newblock In {\em ICLR}, 2016.

\bibitem{Markman2009}
K.~D. Markman, W.~M.~P. Klein, and J.~A. Suhr.
\newblock {\em Handbook of Imagination and Mental Simulation}.
\newblock Taylor \& Francis Group, 2009.

\bibitem{Mathieu2016}
M.~Mathieu, C.~Couprie, and Y.~LeCun.
\newblock Deep multi-scale video prediction beyond mean square error.
\newblock In {\em ICLR}, 2016.

\bibitem{Metz2017}
L.~Metz, B.~Poole, D.~Pfau, and J.~Sohl-Dickstein.
\newblock Unrolled generative adversarial networks.
\newblock In {\em ICLR}, 2017.

\bibitem{Nguyen2016}
A.~Nguyen, A.~Dosovitskiy, J.~Yosinski, T.~Brox, and J.~Clune.
\newblock Synthesizing the preferred inputs for neurons in neural networks via
  deep generator networks.
\newblock In {\em NIPS}, 2016.

\bibitem{Nguyen2017}
A.~Nguyen, J.~Yosinski, Y.~Bengio, A.~Dosovitskiy, and J.~Clune.
\newblock Plug {\&} play generative networks: Conditional iterative generation
  of images in latent space.
\newblock In {\em CVPR}, 2017.

\bibitem{Nguyen2015}
A.~Nguyen, J.~Yosinski, and J.~Clune.
\newblock Deep neural networks are easily fooled: High confidence predictions
  for unrecognizable images.
\newblock In {\em CVPR}, 2015.

\bibitem{Odena2016}
A.~Odena, V.~Dumoulin, and C.~Olah.
\newblock Deconvolution and checkerboard artifacts.
\newblock {\em Distill}, 2016.
\newblock http://distill.pub/2016/deconv-checkerboard.

\bibitem{Oh2015}
J.~Oh, X.~Guo, H.~Lee, R.~L. Lewis, and S.~P. Singh.
\newblock Action-conditional video prediction using deep networks in {Atari}
  games.
\newblock In {\em NIPS}, 2015.

\bibitem{Pathak2016}
D.~Pathak, P.~Kr{\"{a}}henb{\"{u}}hl, J.~Donahue, T.~Darrell, and A.~A. Efros.
\newblock Context encoders: Feature learning by inpainting.
\newblock In {\em CVPR}, 2016.

\bibitem{Pharr2016}
M.~Pharr, W.~Jakob, and G.~Humphreys.
\newblock {\em Physically Based Rendering: From Theory to Implementation}.
\newblock Morgan Kaufmann, 3rd edition, 2016.

\bibitem{PortillaSimoncelli2000}
J.~Portilla and E.~P. Simoncelli.
\newblock A parametric texture model based on joint statistics of complex
  wavelet coefficients.
\newblock {\em IJCV}, 40(1), 2000.

\bibitem{Radford2016}
A.~Radford, L.~Metz, and S.~Chintala.
\newblock Unsupervised representation learning with deep convolutional
  generative adversarial networks.
\newblock In {\em ICLR}, 2016.

\bibitem{Reed2016:NIPS}
S.~E. Reed, Z.~Akata, S.~Mohan, S.~Tenka, B.~Schiele, and H.~Lee.
\newblock Learning what and where to draw.
\newblock In {\em NIPS}, 2016.

\bibitem{Reed2016:ICML}
S.~E. Reed, Z.~Akata, X.~Yan, L.~Logeswaran, B.~Schiele, and H.~Lee.
\newblock Generative adversarial text to image synthesis.
\newblock In {\em ICML}, 2016.

\bibitem{Ronneberger2015}
O.~Ronneberger, P.~Fischer, and T.~Brox.
\newblock {U-Net}: Convolutional networks for biomedical image segmentation.
\newblock In {\em MICCAI}, 2015.

\bibitem{Salimans2016}
T.~Salimans, I.~J. Goodfellow, W.~Zaremba, V.~Cheung, A.~Radford, and X.~Chen.
\newblock Improved techniques for training {GANs}.
\newblock In {\em NIPS}, 2016.

\bibitem{Sayood2012}
K.~Sayood.
\newblock {\em Introduction to Data Compression}.
\newblock Morgan Kaufmann, 2012.

\bibitem{Shan2013}
Q.~Shan, R.~Adams, B.~Curless, Y.~Furukawa, and S.~M. Seitz.
\newblock The visual {Turing} test for scene reconstruction.
\newblock In {\em 3DV}, 2013.

\bibitem{Silberman2012}
N.~Silberman, D.~Hoiem, P.~Kohli, and R.~Fergus.
\newblock Indoor segmentation and support inference from {RGBD} images.
\newblock In {\em ECCV}, 2012.

\bibitem{SimonyanZisserman2015}
K.~Simonyan and A.~Zisserman.
\newblock Very deep convolutional networks for large-scale image recognition.
\newblock In {\em ICLR}, 2015.

\bibitem{Srivastava2015}
N.~Srivastava, E.~Mansimov, and R.~Salakhutdinov.
\newblock Unsupervised learning of video representations using {LSTMs}.
\newblock In {\em ICML}, 2015.

\bibitem{Tatarchenko2016}
M.~Tatarchenko, A.~Dosovitskiy, and T.~Brox.
\newblock Multi-view {3D} models from single images with a convolutional
  network.
\newblock In {\em ECCV}, 2016.

\bibitem{Vondrick2016}
C.~Vondrick, H.~Pirsiavash, and A.~Torralba.
\newblock Generating videos with scene dynamics.
\newblock In {\em NIPS}, 2016.

\bibitem{WangGupta2016}
X.~Wang and A.~Gupta.
\newblock Generative image modeling using style and structure adversarial
  networks.
\newblock In {\em ECCV}, 2016.

\bibitem{Xue2016}
T.~Xue, J.~Wu, K.~L. Bouman, and B.~Freeman.
\newblock Visual dynamics: Probabilistic future frame synthesis via cross
  convolutional networks.
\newblock In {\em NIPS}, 2016.

\bibitem{Yan2016:ECCV}
X.~Yan, J.~Yang, K.~Sohn, and H.~Lee.
\newblock {Attribute2Image}: Conditional image generation from visual
  attributes.
\newblock In {\em ECCV}, 2016.

\bibitem{YuKoltun2016}
F.~Yu and V.~Koltun.
\newblock Multi-scale context aggregation by dilated convolutions.
\newblock In {\em ICLR}, 2016.

\bibitem{Zhou2016}
T.~Zhou, S.~Tulsiani, W.~Sun, J.~Malik, and A.~A. Efros.
\newblock View synthesis by appearance flow.
\newblock In {\em ECCV}, 2016.

\bibitem{Zhu2016}
J.-Y. Zhu, P.~Kr{\"{a}}henb{\"{u}}hl, E.~Shechtman, and A.~A. Efros.
\newblock Generative visual manipulation on the natural image manifold.
\newblock In {\em ECCV}, 2016.

\end{thebibliography}


\begin{thebibliography}{1}\itemsep=-1pt

\bibitem{Cordts2016}
M.~Cordts, M.~Omran, S.~Ramos, T.~Rehfeld, M.~Enzweiler, R.~Benenson,
  U.~Franke, S.~Roth, and B.~Schiele.
\newblock The {Cityscapes} dataset for semantic urban scene understanding.
\newblock In {\em CVPR}, 2016.

\bibitem{Isola2017}
P.~Isola, J.~Zhu, T.~Zhou, and A.~A. Efros.
\newblock Image-to-image translation with conditional adversarial networks.
\newblock {\em arXiv:1611.07004}, 2016.

\bibitem{Long2015}
J.~Long, E.~Shelhamer, and T.~Darrell.
\newblock Fully convolutional networks for semantic segmentation.
\newblock In {\em CVPR}, 2015.

\bibitem{Richter2016}
S.~R. Richter, V.~Vineet, S.~Roth, and V.~Koltun.
\newblock Playing for data: Ground truth from computer games.
\newblock In {\em ECCV}, 2016.

\bibitem{Ronneberger2015}
O.~Ronneberger, P.~Fischer, and T.~Brox.
\newblock {U-Net}: Convolutional networks for biomedical image segmentation.
\newblock In {\em MICCAI}, 2015.

\end{thebibliography}
